\documentclass[runningheads]{llncs}
\usepackage{graphicx}

\usepackage{tikz}
\usepackage{comment}
\usepackage{amsmath,amssymb} 
\usepackage{color}

\usepackage{xcolor}
\usepackage{amssymb}
\usepackage{pifont}
\usepackage{listings}
\usepackage{url}
\usepackage{graphicx}
\usepackage{wrapfig}
\usepackage{booktabs}
\usepackage{makecell}
\usepackage{colortbl}
\usepackage{multirow}
\usepackage{fixltx2e}
\usepackage{subcaption}
\usepackage{amsmath}
\usepackage{etoolbox}
\usepackage{multicol}
\usepackage{refcount}
\usepackage{marvosym}
\usepackage{ifsym}
\usepackage{hyperref}
\definecolor{mygray}{gray}{0.6}

\usepackage[accsupp]{axessibility}  

\definecolor{blue}{rgb}{0.31, 0.78, 0.47}
\definecolor{red}{rgb}{0.8, 0.0, 0.0}

\hypersetup{
    colorlinks=true,
    linkcolor=blue,
    filecolor=magenta,      
    urlcolor=cyan,
    pdftitle={TokenMix},
    pdfpagemode=FullScreen,
    }

\begin{document}
\pagestyle{headings}
\mainmatter
\def\ECCVSubNumber{4037}  

\title{TokenMix: Rethinking Image Mixing for Data Augmentation in Vision Transformers} 

\titlerunning{TokenMix}
%
\author{Jihao Liu\inst{1,2} \and Boxiao Liu\inst{3} \and Hang Zhou\inst{1} \and Hongsheng Li\inst{1}\textsuperscript{\Letter} \and Yu Liu\inst{2}\textsuperscript{\Letter}}

\footnotetext{\textsuperscript{\Letter} Corresponding author.}

\authorrunning{Liu et al.}
%
\institute{CUHK, MMLab \and SenseTime Research \and SKLP, Institute of Computing Technology, CAS}

\maketitle

\begin{abstract}
   CutMix is a popular augmentation technique commonly used for training modern convolutional and transformer vision networks. It was originally designed to encourage Convolution Neural Networks (CNNs) to focus more on an image's global context instead of local information, which greatly improves the performance of CNNs. However, we found it to have limited benefits for transformer-based architectures that naturally have a global receptive field. In this paper, we propose a novel data augmentation technique TokenMix to improve the performance of vision transformers. TokenMix mixes two images at token level via partitioning the mixing region into multiple separated parts. Besides, we show that the mixed learning target in CutMix, a linear combination of a pair of the ground truth labels, might be inaccurate and sometimes counter-intuitive. To obtain a more suitable target, we propose to assign the target score according to the content-based neural activation maps of the two images from a pre-trained teacher model, which does not need to have high performance. With plenty of experiments on various vision transformer architectures, we show that our proposed TokenMix helps vision transformers focus on the foreground area to infer the classes and enhances their robustness to occlusion, with consistent performance gains. Notably, we improve DeiT-T/S/B with +1\% ImageNet top-1 accuracy. Besides, TokenMix enjoys longer training, which achieves 81.2\% top-1 accuracy on ImageNet with DeiT-S trained for 400 epochs. Code is available at \url{https://github.com/Sense-X/TokenMix}.
   \noindent
\end{abstract}

\section{Introduction}
\label{sec:intro}

Deep neural networks dominate the learning of visual representations and show effectiveness on various downstream tasks, including image classification \cite{imagenet,vit}, object detection \cite{coco}, semantic
segmentation \cite{ade20k}, etc. To further improve the performance, various data augmentation strategies were introduced, including hand-crafted \cite{zhang2017mixup,cutmix} and automatically searched ones \cite{autoaug,randaugment}.
Recently, data augmentation based on mixing multiple images into single ones shows impressive performances on various vision tasks. The labels of such ``mixed'' images are created based on their original labels.
Mixup \cite{zhang2017mixup} for the first time attempted to generate mixed training samples via linear combinations of pairs of samples. CutMix \cite{cutmix} proposed to mix pairs of samples on region level, which replaces a random local rectangular area in a source image with the contents of the corresponding area in a target image. In addition, a series of works attempted to improve CutMix with more complicated strategies on choosing rectangular sizes and locations to be used for mixing \cite{attentivemix,puzzlemix,comix}.

\begin{figure}[t]
    \centering
    \includegraphics[width=0.5\linewidth]{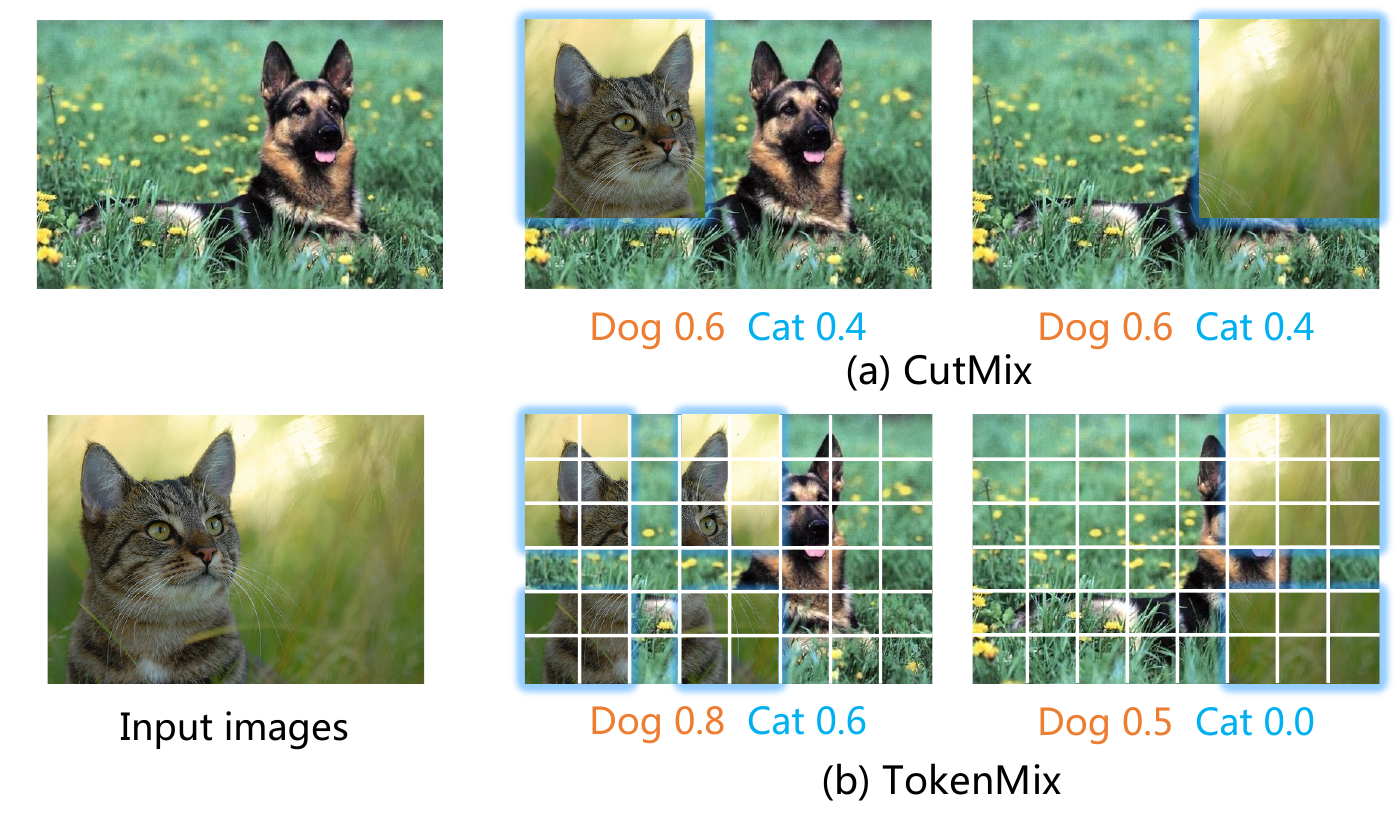}
    \caption{\textbf{TokenMix} and CutMix. TokenMix not only mixes images at token-level to encourage better learning of long-range dependency but also generates more reasonable target scores according to content-based neural activation maps from an (even imperfectly trained) teacher network.}
    \label{fig:example}
\end{figure}

In general, CutMix and its variants use the region-level cut-and-paste mixing technique to enforce Convolution Neural Networks (CNNs) to pay more attention to the image's global context instead of just local information.
While the CutMix augmentation can also be used for training vision transformers \cite{vit,deit}, the region-level mixing strategy becomes less effective.

We revisit the design of CutMix augmentation and argue that it is a sub-optimal strategy for transformer-based architectures. On the one hand, the region-level mixing in CutMix cuts a rectangular area in a source image and mixes the contents into a target image. As CNNs are primarily designed to encode local image contents, the region-level mixing of CutMix can effectively prevent CNNs from over-focusing on local context. However, for transformer-based architectures that naturally have global receptive fields from the first layer, the region-level mixing is less beneficial. On the other hand, CutMix assigns a mixed label for the augmented image according to only the cropped area ratio between the source and target images, regardless of their cropped contents. However, the cut region and location of CutMix are randomly chosen, and the same label is assigned no matter whether the cut contents are foreground or background, which inevitably introduces label noise to the learning targets and causes unstable training (see Fig. \ref{fig:example}(b)). There are recent works trying to mitigate this problem by attentively choosing the salient area for cutting \cite{attentivemix,saliencymix} or using alternate optimization to determine the cutting region \cite{puzzlemix,comix}. However, the label noise problem is still under-explored as the salient areas might not correctly correspond to the foreground regions.

In this paper, we propose TokenMix, a token-level augmentation technique that can be well applied to training various transformer-based architectures. In contrast to previous approaches, TokenMix directly mixes two images at the token level to promote the interactions of the input tokens and generates a more reasonable target with considering the images' semantic information.
First, to train transformers for better encoding long-range dependency, we directly cut at token level and allow the cut region to be separated into multiple isolated parts. As a result, the cut area can be distributed all over the image, as shown in Figure~\ref{fig:example} (b). The token-level mixing encourages the transformer to better encode long-range dependency to correctly classify the 
mixed images with the augmented tokens inside. Instead of relying on alternate optimization or an extra network to determine which region to mix, all the mixing tokens in TokenMix are randomly determined as blocks, which is easier to implement with a small number of hyper-parameters. 

Besides, previous methods usually assign a mixed target to the augmented image, which equals the linear combination of the ground truth labels of the source and target images. The linear combination ratio of the labels is determined as the area ratio between the cutting region of the source image and the total size of the target image.
We found that such target scores can be highly inaccurate. As shown in Fig.~\ref{fig:example} (a), the same target is assigned to both cases even the mixing area has significantly different semantic meanings.

Following the spirit of distillation, we propose to assign the target score to an augmented target image according to the content-based neural activation maps of the two mixing images. Specifically, we first obtain the neural activation maps of both the source and target images with a pre-trained neural network, which does not need to be perfectly trained. The scores of two mixing regions are calculated as the summation of the spatially normalized neural activation maps, which are combined as the final target. Our intuition is that the neural activation map of even a partially trained classification network can better localize some part of an object \cite{cam,vistransformer} than using naive score averaging. After spatial normalization of the neural activation map, the regions with rich semantic information would be assigned high scores and low scores would be assigned for other regions, leading to more robust targets. 
The neural activation maps are generated offline, so the extra training overhead introduced is negligible (+0.8\%). In contrast, the distillation method used in DeiT \cite{deit} relies on the online inference of a teacher network to generate target scores from augmented images, which cannot generate target scores offline and therefore nearly doubles the training time.
Although ReLabel \cite{relabel} and TokenLabeling \cite{tokenlabeling} also explored utilizing neural activation maps to generate training supervisions, their approaches use the patch-level activations as supervisions and is prone to suffer more from inaccurate activation maps of mix-based augmentations. In contrast, our proposed method sums up the activations from the cut regions as the image-level target scores and is less likely to be affected by individual tokens' incorrect activations. Experiments on combining our token cutting strategy and ReLabel or TokenLabeling validate our scoring strategy.
We show that the resulted targets of our approach are more reasonable, which improve the performance and stabilize the training of not only our proposed TokenMix and also the original CutMix. 
Replacing the way to generate target scores in CutMix with our approach, we obtain a +0.7\% top-1 accuracy gain on ImageNet with DeiT-S.
In addition, as the generated target scores are more learning-friendly, 
we show that our approach enjoys longer training. Specifically, we achieve 81.2\% top-1 accuracy on ImageNet with DeiT-S when training for 400 epochs. 

In summary, our contributions are as follows:
\begin{itemize}
    \item We propose TokenMix, a token-level augmentation technique that generalizes well across various transformer-based architectures.
    
    \item We propose to assign the target scores of the mixed images with content-based neural activation maps, which can benefit both TokenMix and CutMix augmentations.
    
    \item Experimental results show that TokenMix promotes transformer's capability on encoding image contents and robustness to the occlusions. We improve DeiT-S from 79.8\% to 80.8\% top-1 accuracy on ImageNet.
\end{itemize}

\section{Related Works}
\label{sec:related}
\noindent\textbf{Cutting-based data augmentation.}
The motivation behind cutting-based methods \cite{Devries2017ImprovedRO,Zhong2020RandomED,Singh2018HideandSeekAD,Chen2020GridMaskDA} is to make a network learn informative representations from the entire image. By masking some areas from the input image, it can alleviate the issue of overfitting and improve the occlusion robustness \cite{Devries2017ImprovedRO}. 
Cutout \cite{Devries2017ImprovedRO} is a pioneer of this idea, and proposes to randomly select a square patch of an image and set the inputs within as some consistent. The shape and size of the masked patch are manually designed.
Random-erasing \cite{Zhong2020RandomED} works in a similar way with Cutout, but introduces more randomness into the augmentation. In every iteration, the erasing operation is performed under a probability, and the size and aspect ratio is randomly selected with predefined limits.
Hide-and-seek \cite{Singh2018HideandSeekAD} differs from the previous two methods in the number of masked patches. It divides an image into grids and masked each grid randomly and independently.

\noindent\textbf{Mixing-based data augmentation.}
Mixing-based data augmentations are another popular regularization method to help the optimization of deep neural networks~\cite{zhang2017mixup,Verma2019ManifoldMB,comix,Hendrycks2020AugMixAS,alignmixup}. 
Mixup \cite{zhang2017mixup} proposes to mix the RGB values of two randomly selected images according to a mixing factor, which is drawn from a beta distribution. The target for the mixed image is also a linear combination of the targets of original images.
Manifold Mixup \cite{Verma2019ManifoldMB} extends the mixed information from input images to intermediate feature maps of a network.
Co-Mixup \cite{comix} and Puzzle Mix \cite{puzzlemix} consider the mixing process as an optimization problem, and propose to maximize the saliency in the mixed images.
AugMix \cite{Hendrycks2020AugMixAS} generates mixed images from the original image and its transformed ones. AlignMixup~\cite{alignmixup} proposes to geometrically align two images in the feature space and then perform feature mixing in the deep layers. 

\noindent\textbf{Joint of cutting and mixing.}
One issue of cutting-based augmentation is the information in the cut area is lost, so recent researches \cite{cutmix,Takahashi2020DataAU,Qin2020ResizeMixMD,Chen2021TransMixAT} propose to combine cutting and mixing together to achieve better performance.
As introduced in CutMix, a patch is replaced with that from another image instead of being deleted. Like Mixup, the target for the mixed image is computed as the proportion of the replaced area.
Attentive CutMix \cite{attentivemix} points out that the randomly selected patches may contain only background regions, and proposes to replace attentive regions identified by a pre-trained network.
RICAP \cite{Takahashi2020DataAU} introduces another way of stitching four rectangle patches from different images into one new image. The target of the new image is also determined according to the area of different patches.
ResizeMix \cite{Qin2020ResizeMixMD} argues the traditional cut-and-paste operation may lead to an unreasonable target when only the background part of one image is mixed. It solves this issue by using a resize-and-paste process. 
In this paper, we revisit the CutMix method for vision transformer and find that CutMix under-explores the ability of vision transformer to model long-range interaction and the assigned target is non-optimal. We further introduce our TokenMix augmentation with novel ways to select mixed parts and generate learning targets.

\section{Method}
\label{sec:method}
In this section, we first revisit the general process of CutMix \cite{cutmix} and show the limitations of applying CutMix to transformers. We then present our proposed TokenMix, which conducts image augmentation via mixing images at token-level and assigns target scores with neural activation maps.

\begin{figure}[t]
    \centering
    \includegraphics[width=0.5\linewidth]{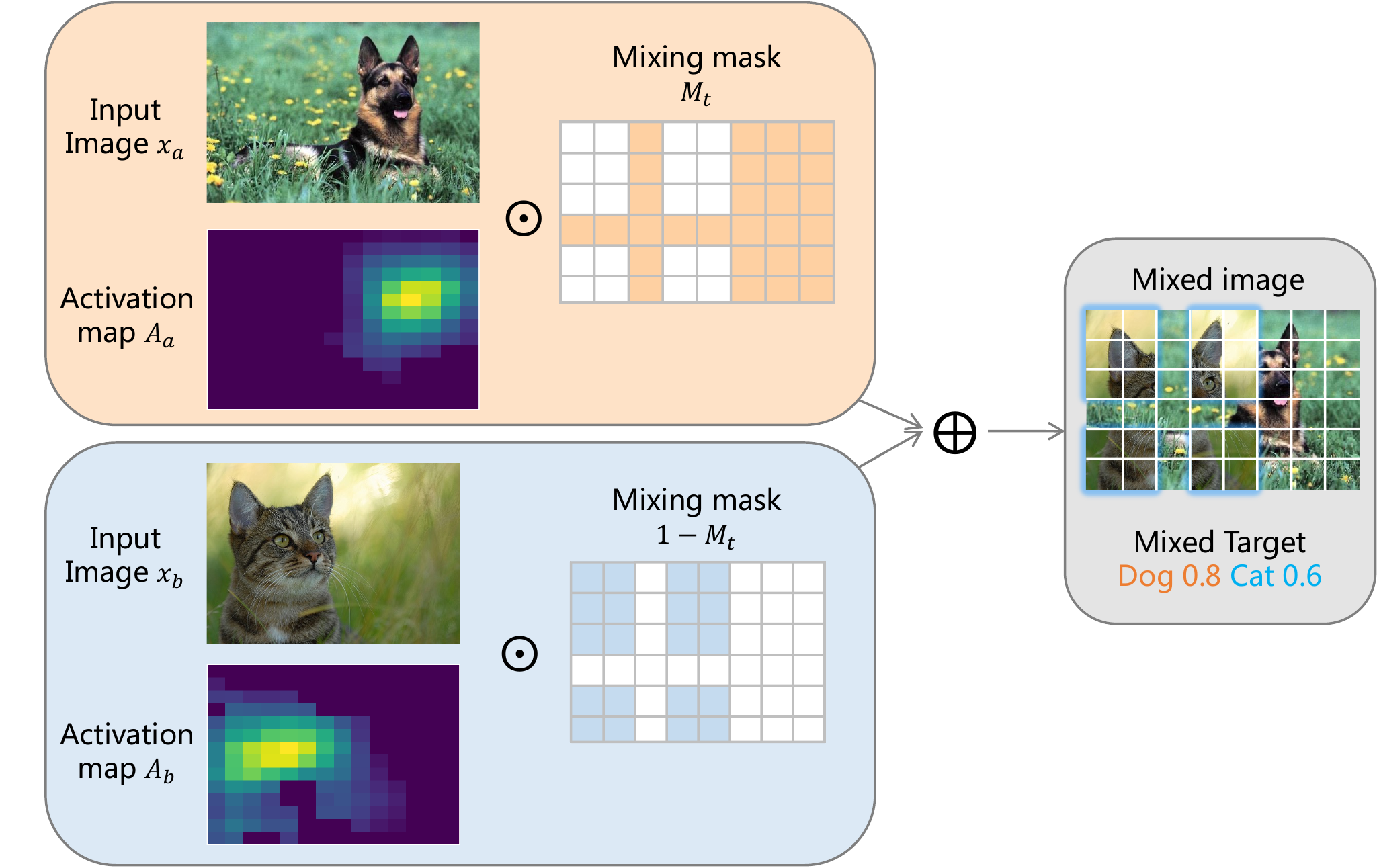}
    \caption{\textbf{The overall pipeline of TokenMix.} TokenMix partitions the mask region into multiple separated parts. The target score of the mixed image is calculated according to the neural activation maps of the two input images.}
    \label{fig:tokenmix}
\end{figure}

\subsection{Revisiting CutMix Augmentation}
\label{sec:revisit}
To enhance the localization ability of CNNs, CutMix \cite{cutmix} proposed to mix pairs of samples with a random rectangular binary mask. Let 
$ x \in \mathbb{R}^{H \times W \times C} $ and $y$ denote a training image and its label, respectively. Given a pair of training samples $(x_a, y_a)$ and $(x_b, y_b)$, CutMix generates a new training sample $(\Tilde{x}, \Tilde{y})$ as follows:
\begin{equation}
    \begin{array}{l}
        \Tilde{x} = M \odot x_a + (1 - M) \odot x_b, \\
        \Tilde{y} = \lambda y_a + (1 - \lambda) y_b,
    \end{array}
    \label{equ:cutmix}
\end{equation}
where $M\in{\{0, 1\}}^{H \times W}$ denotes the rectangular mask that decides where to drop out and fill in the contents of the two images, $\odot$ denotes element-wise multiplication, and $\lambda$ is sampled from a beta distribution Beta$(\alpha, \alpha)$. The binary mask $M$ is a randomly sampled rectangle, which guarantees $\frac{\sum M}{HW} = \lambda$.
Similar to Mixup \cite{zhang2017mixup}, CutMix assigns a mixed target for the generated image as a linear combination of $y_a$ and $y_b$. 

We argue that the region-level mixing in CutMix might not be suitable for transformer-based architectures.
As CNNs are primarily designed to encode local image contents, training with CutMix effectively prevents CNNs from over-focusing on local context. However, transformer-based architectures might be less benefited from CutMix as all of its layers have global receptive fields. In addition, the label of the mixed image is a linear combination of $y_a$ and $y_b$ with mixing ratio $\lambda$ being estimated only according to the size of the mask, which might be inappropriate in many cases as shown in Figure \ref{fig:example} (b). Although there were recent methods on attempting to improve CutMix by choosing the salience regions to maximize the saliency in the mixed images \cite{comix,puzzlemix,attentivemix,saliencymix}, the salient areas might not correctly correspond to the target class \cite{imagenetdone}, and the label noise problem is still serious.

\subsection{TokenMix}
\label{sec:tokenmix}
In this paper, we propose TokenMix to mix a pair of images to generate a mixed image and learning target. We generate the mask $M$ at token-level to encourage better learning of long-range dependency and assign the target score of the mixed image according to the content-based neural activation maps of the two mixing images, which follows the general spirit of distillation to create more robust targets.

Figure \ref{fig:tokenmix} shows an overview of our proposed TokenMix.
We first partition the input image $x$ into non-overlapping patches $x^p \in \mathbb{R}^{\frac{H}{P} \times \frac{W}{P} \times (P^2\cdot C)}$, which are then linearly projected to visual tokens. We then generate a random mask $M_t \in \mathbb{R}^{\frac{H}{P} \times \frac{W}{P}}$ at token-level according to the mask-out ratio $\lambda$. The mixed new training sample $(\Tilde{x}^p, \Tilde{y})$ is created as follows:
\begin{equation}
    \begin{array}{l}
        \Tilde{x}^p = M_t \odot x^p_a + (1 - M_t) \odot x^p_b, \\
        \Tilde{y} = \sum\limits_{i \in \mathfrak{S}} M_{ti} \odot A_{ai} + \sum\limits_{i \in \mathfrak{S}} (1 - M_{ti}) \odot A_{bi},
    \end{array}
    \label{equ:tokenmix}
\end{equation}
where $\mathfrak{S}$ indicates the set of all tokens, $\odot$ denotes element-wise multiplication, $M_{ti}$ denotes the $i$-th token of the mask $M_t$, $A_{ai}$ and $A_{bi}$ are the $i$-th token of the spatially normalized neural activation maps of $x_a$ and $x_b$ respectively. The neural activation maps are generated with pre-trained networks' last layer before the classification head \cite{tokenlabeling,relabel}. 

Instead of masking a whole rectangular area, we partition the mask area into multiple separated parts. For each part, we randomly choose the number of masked tokens and the aspect ratio \cite{beit,cutmix}. We set the minimum number of tokens to 14 and log-uniformly sample the aspect ratio in the range of $[0.3, \frac{1}{0.3}]$.
We repeatedly mask a part of the image until the total number of masked tokens reaches the pre-defined ratio $\lambda\frac{HW}{P^2}$. Instead of sampling $\lambda$ from a  beta distribution, we set $\lambda$ to 0.5 unless otherwise specified. Our intuition is that the distributed masking regions are easier to recognize compared with masking a whole rectangular area.
For investigation, we also introduce a uniformly random version, where each masking part is only a single token. While totally random mixing is harmful to the performance of CNNs, we show that transformers are still benefited from the simplified version.

To solve the issue of inaccurate target scores generated by CutMix, we propose to set the target score with the content-based neural activation maps of the two mixing images, generated by a pre-trained teacher network. Our intuition is that not all regions correspond to the foreground object. Concretely, the regions with rich semantic information would have a bigger impact on the target score than other regions. Inspired by the distillation technique that sets the target score of an image by a teacher network, we extend the design to set the target score by combining a teacher network's neural activation maps of the two mixing images.
As shown in Figure \ref{fig:tokenmix}, the target scores of two mixing regions are calculated as the summation of spatially normalized neural activation maps within the mask for $x_a$ or outside the mask for $x_b$. We then combine the two target scores as the final target of the mixed image. 

Compared to previous arts \cite{cutmix,attentivemix,puzzlemix,comix}, our proposed TokenMix has two main advantages: 1) We explicitly encourage the transformer to better encode long-range dependency to correctly classify the image with the other image mixed inside. We show that our approach can lead to consistent accuracy gain when used in various vision transformers, and also enhances the occlusion robustness of the transformers. 2) The target label of the mixed image that is generated with content-based neural activation map is more robust than those of previous approaches, which takes advantage of the distillation technique. Besides, we show that our approach promotes transformers to better localize the discriminate regions, with attention weights.

\section{Experiment}
\label{sec:experiment}

\subsection{Datasets}
\label{sec:setup}

We use ImageNet-1K \cite{imagenet} dataset to demonstrate the effectiveness of our method. The dataset contains 1.2 million images for training and 50K for validation. The top-1 accuracy is reported as the evaluation metric.

We also use ADE20K \cite{ade20k} to verify the transferability of our TokenMix pre-trained models. ADE20K is a widely-used semantic segmentation dataset, covering 150 semantic categories. The dataset has 25K images in total, with 20K for training, 2K for validation, and another 3K for testing.

\subsection{Implementation Details}
We evaluate our method on several recent vision transformer architectures, including DeiT \cite{deit}, CaiT \cite{cait}, PVT \cite{pvt} and Swin Transformer \cite{pvt}. We also test TokenMix on ResNet \cite{resnet}, which is representative of convolution models, as comparison. We follow the training recipe of DeiT \cite{deit}. The batch size is set to 1024. We use AdamW \cite{adam,adamw} as the optimizer and set the learning rate as 0.001 with 5 warm-up epochs. The learning rate is decayed following a cosine scheduler down to $10^{-6}$. Without other specification, we train the models for 300 epochs. Rand Augment \cite{randaugment} and Mixup \cite{zhang2017mixup} are both used by default. Following \cite{deit}, we switch TokenMix and Mixup with the probability of 0.5. For training architectures with smaller model sizes, e.g., DeiT-T \cite{deit}, PVT-T \cite{pvt}, or CaiT-XXS \cite{cait}, we sample the $\lambda$ in Equation \ref{equ:tokenmix} from a beta distribution Beta$(1.0, 1.0)$. We use binary cross-entropy (BCE) loss instead of the typical cross-entropy (CE) loss by default following \cite{timmresnet,imagenetdone}, as the mixed images are more likely to contain multiple labels. To generate the neural activation maps, we defautly use NFNet-F6 \cite{nfnet} following \cite{tokenlabeling}. 

For transferring to the ADE20K dataset, we follow the setting in BEiT \cite{beit}, and fine-tune for 160K steps with Adam \cite{adam} optimizer. The detailed hyperparameters are described in supplementary materials.

\section{Main Results}
\label{sec:result}

\begin{table}[t]
  \centering
  \caption{ImageNet classification performances based on various transformer-based architectures. TokenMix consistently improves DeiT for $\sim$ 1\% top-1 accuracy with nearly no extra training overhead.}
  \resizebox{0.65\linewidth}{!}{
    \begin{tabular}{l|cc|c|c}
    \toprule
    Model & \makecell{\#FLOPs (G) } & \makecell{\#Params (M)} & CutMix & TokenMix \\
    \midrule
     
    DeiT-T \cite{deit} & 1.3   & 5.7  & 72.2 & \textbf{73.2 \textcolor{blue}{(+1.0)}} \\
    PVT-T \cite{pvt} &  1.9   & 13.2  & 75.1 & \textbf{75.6 \textcolor{blue}{(+0.5)}} \\
    CaiT-XXS-24 \cite{cait} & 2.5   & 9.5  & 77.6 & \textbf{78.0 \textcolor{blue}{(+0.4)}} \\
    DeiT-S \cite{deit} & 4.6   & 22.1  & 79.8 & \textbf{80.8 \textcolor{blue}{(+1.0)}} \\
    Swin-T \cite{swin} & 4.5   & 29  & 81.2 & \textbf{81.6 \textcolor{blue}{(+0.4)}} \\
    DeiT-B \cite{deit} & 17.6  & 86.6 & 81.8 & \textbf{82.9 \textcolor{blue}{(+1.1)}} \\
    
    \bottomrule
    \end{tabular}%
    }
  \label{tab:main_results}
\end{table}%

\subsection{ImageNet Results}
We report the results on ImageNet-1K dataset with our TokenMix. As shown in Table \ref{tab:main_results}, TokenMix consistently improves CutMix on various transformer-based architectures, i.e., DeiT \cite{deit}, PVT \cite{pvt}, CaiT \cite{cait}, and Swin Transformer \cite{swin}. Specifically, TokenMix outperforms CutMix \cite{cutmix} by +1\% for DeiT, across DeiT-T to DeiT-B. We also improve popular hierarchical transformer architectures Swin-T and PVT-T for +0.4\% and +0.5\%, respectively.
All the results demonstrate the effectiveness and generalization of the proposed TokenMix. 

Our proposed TokenMix consists of two parts, i.e., token-level mixing and label refinement. We decouple the two parts and then compare them with the previous methods by fixing one part. In Table \ref{tab:comp_label}, we compare TokenMix to ReLabel \cite{relabel} and TokenLabeling \cite{tokenlabeling} with the same data augmentation method. The two methods utilize pixel-level supervision, but our TokenMix summarizes neural activations to create image-level target scores and is, therefore, more robust to individual pixel-level errors. Note that we use the same teacher network, i.e., NFNet-F6, to generate the offline targets.
As shown in Table \ref{tab:comp_label}, TokenMix outperforms both ReLabel (+0.5\%) and TokenLabeling (+0.3\%) with the same training cost. 
We further compare TokenMix to previous mixing-based augmentation methods in Table \ref{tab:comp_mix}.  For a more fair comparison, we only use the labels from ImageNet.
As shown in Table \ref{tab:comp_mix}, TokenMix have performance advantages compared to other approaches. We see that the methods that introduce more foreground regions fail to improve CutMix on Vision Transformer. In contrast, our proposed TokenMix improves CutMix for +0.5\% accuracy. 

\begin{table}[t]
    \centering
    \begin{minipage}[t]{0.57\linewidth}
    \captionof{table}{Comparisons with ReLabel and TokenLabeling with DeiT-T on ImageNet. GPU time refers to the increase of training time.}
    \centering
    \resizebox{0.67\linewidth}{!}{
        \begin{tabular}{cccc}
            \toprule
            Augmentation & Supervision & \makecell{Top-1 \\ Acc.} & \makecell{GPU \\ Time} \\
            \midrule
            CutMix   & ImageNet & 72.2 & +0.0\% \\
            TokenMix & ImageNet & \textbf{72.7} & +0.0\% \\
            \midrule
            TokenMix & ReLabel & 72.7 & +0.8\% \\
            TokenMix & TokenLabeling & 72.9 & +0.8\% \\
            TokenMix & TokenMix & \textbf{73.2} & +0.8\% \\
            \bottomrule
        \end{tabular}
    }
    \label{tab:comp_label}
    \end{minipage}
    \centering
    \begin{minipage}[t]{0.4\linewidth}
        \centering
        \captionof{table}{Comparisons with previous mixing methods with DeiT-T on ImageNet.}
        \resizebox{0.55\linewidth}{!}{
            \begin{tabular}{ccc}
                \toprule
                Augmentation & \makecell{Top-1 \\ Acc.} \\
                \midrule
                CutMix \cite{cutmix} & 72.2 \\
                Co-Mix \cite{comix} & 72.2 \\
                SaliencyMix \cite{saliencymix} & 71.8 \\
                Puzzle-Mix \cite{puzzlemix} & 72.3 \\
                TokenMix & \textbf{72.7} \\
                \bottomrule
            \end{tabular}
        }
        \label{tab:comp_mix}
    \end{minipage}
\end{table}

\begin{table}[t]
    \centering
    \caption{Transferring the pre-trained models to downstream semantic segmentation task on ADE20K dataset. TL and RL denote TokenLabeling and ReLabel respectively. \textcolor{mygray}{\ding{51}}+RL/TL represents row 3/4 in Table \ref{tab:comp_label}.}
    \resizebox{0.55\linewidth}{!}{
    \begin{tabular}{cccccc}
        \toprule
        Model & TokenMix & mIoU(\%) & mAcc(\%) & \makecell{+ms \\ mIoU(\%)} & \makecell{+ms \\ mAcc(\%)} \\
        \midrule
        \multirow{4}{*}{DeiT-T}  & \textcolor{mygray}{\ding{55}} & 36.4 & 46.7 & 37.5 & 47.1 \\
         & \textcolor{mygray}{\ding{51}}+RL & 36.6 & 47.0 & 38.1 & 47.9 \\
         & \textcolor{mygray}{\ding{51}}+TL & 36.9 & 47.1 & 38.3 & 48.1 \\
           & \ding{51} & \textbf{37.1} & \textbf{47.5} & \textbf{38.6} & \textbf{48.2} \\
        \midrule
        \multirow{2}{*}{DeiT-S}  & \textcolor{mygray}{\ding{55}} & 42.3 & 52.8 & 43.7 & 53.8 \\
           & \ding{51} & \textbf{44.5} & \textbf{55.0} & \textbf{45.9} & \textbf{56.1} \\
        \midrule
        \multirow{2}{*}{DeiT-B}  & \textcolor{mygray}{\ding{55}} & 46.3 & 56.5 & 47.7 & 57.6 \\
           & \ding{51} & \textbf{46.8} & \textbf{56.9} & \textbf{48.2} & \textbf{58.1} \\
        \bottomrule
    \end{tabular}
    }
    \label{tab:transfer}
\end{table}

\subsection{Transfer to Downstream Task}

Pre-training on ImageNet-1K then finetuning to the downstream tasks is a common practice for many visual recognition tasks. It is important to verify whether the better pre-train with TokenMix can boost the performance on the downstream task. To do so, we transfer our TokenMix pre-trained models to the semantic segmentation task and compare them with regular pre-train. Note that TokenMix does not introduce extra computation overhead in the transfer stage. As shown in Table \ref{tab:transfer}, we find that better pre-train from TokenMix consistently improves the segmentation performance on the ADE20K dataset. Notably, we improve DeiT-T for +0.7\% mIoU, DeiT-S for +2.2\% mIoU, DeiT-B for +0.5\% mIoU. We notice that the performance gap becomes even larger (e.g. +1.1\% mIoU for DeiT-T) when using multi-scale testing. All the results demonstrate the transferability of our TokenMix pre-trained models.

\subsection{Main Properties}
\label{sec:properties}

\begin{figure*}[t]
    \centering
    \includegraphics[width=0.8\linewidth]{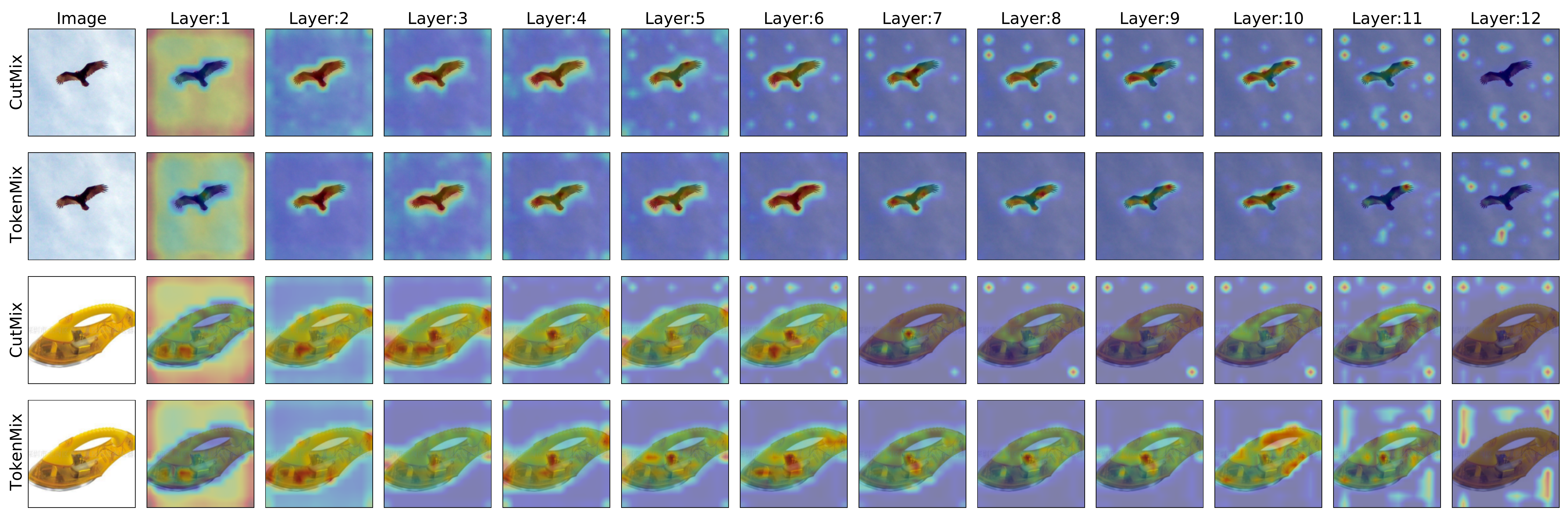}
    \caption{Visualization of the attention maps of the class token in DeiT-S to attend to patch tokens at different layers. Using CutMix distracts the attention to background areas in the several middle layers. In contrast, the proposed TokenMix helps the class token focus more on foreground objects and leads to consistent performance gain.}
    \label{fig:foreground}
\end{figure*}

\begin{figure*}[t]
    \centering
        \includegraphics[width=0.8\textwidth]{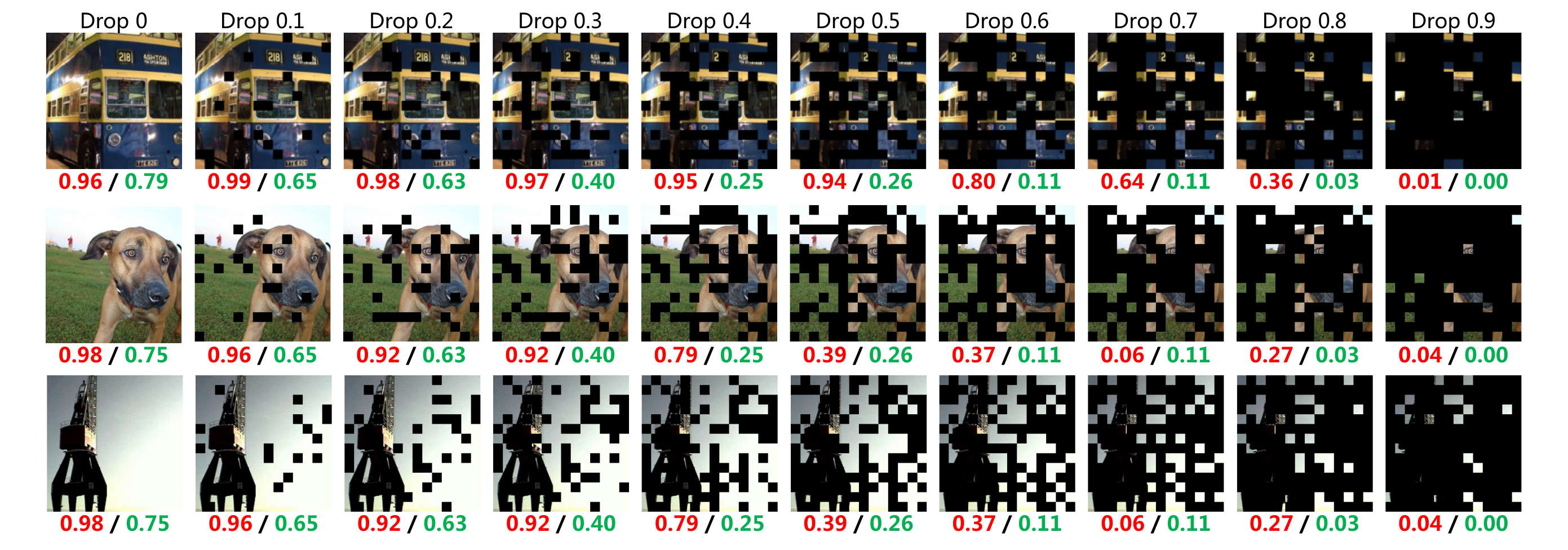}
        \caption{Example images and the predicted confidences under different occlusion ratios. Red scores under the images are predicted by \textcolor{red}{TokenMix}, and green ones by \textcolor{blue}{CutMix}. The model trained with TokenMix holds high confidence when a large number of patches are dropped, while the model trained with CutMix outputs low confidence.}
        \label{fig:occlusion_robustness}
\end{figure*}

Besides the performance gains, we find that our proposed TokenMix improves transformers to be robust to occlusion, and focus more on the foreground area. All the visualization and analysis are conducted on DeiT-S.

\noindent\textbf{TokenMix helps transformers focus on the foreground area.}
As discussed in Section \ref{sec:method}, CutMix assigns targets of the mixed images based on linear combinations of labels of the pairs of mixing images, which might be inaccurate if the foreground region is cut. We find that the inaccurate labels make transformers pay incorrect attention to the input image. As shown in Figure \ref{fig:foreground}, using CutMix distracts the transformer's attention to background areas in several middle layers (layers 5-10). In comparison, TokenMix helps transformers learn to pay more attention to the foreground areas and leads to consistent performance gain.
\noindent\begin{wrapfigure}{r}{0.4\textwidth}
    \centering
    \includegraphics[width=0.9\linewidth]{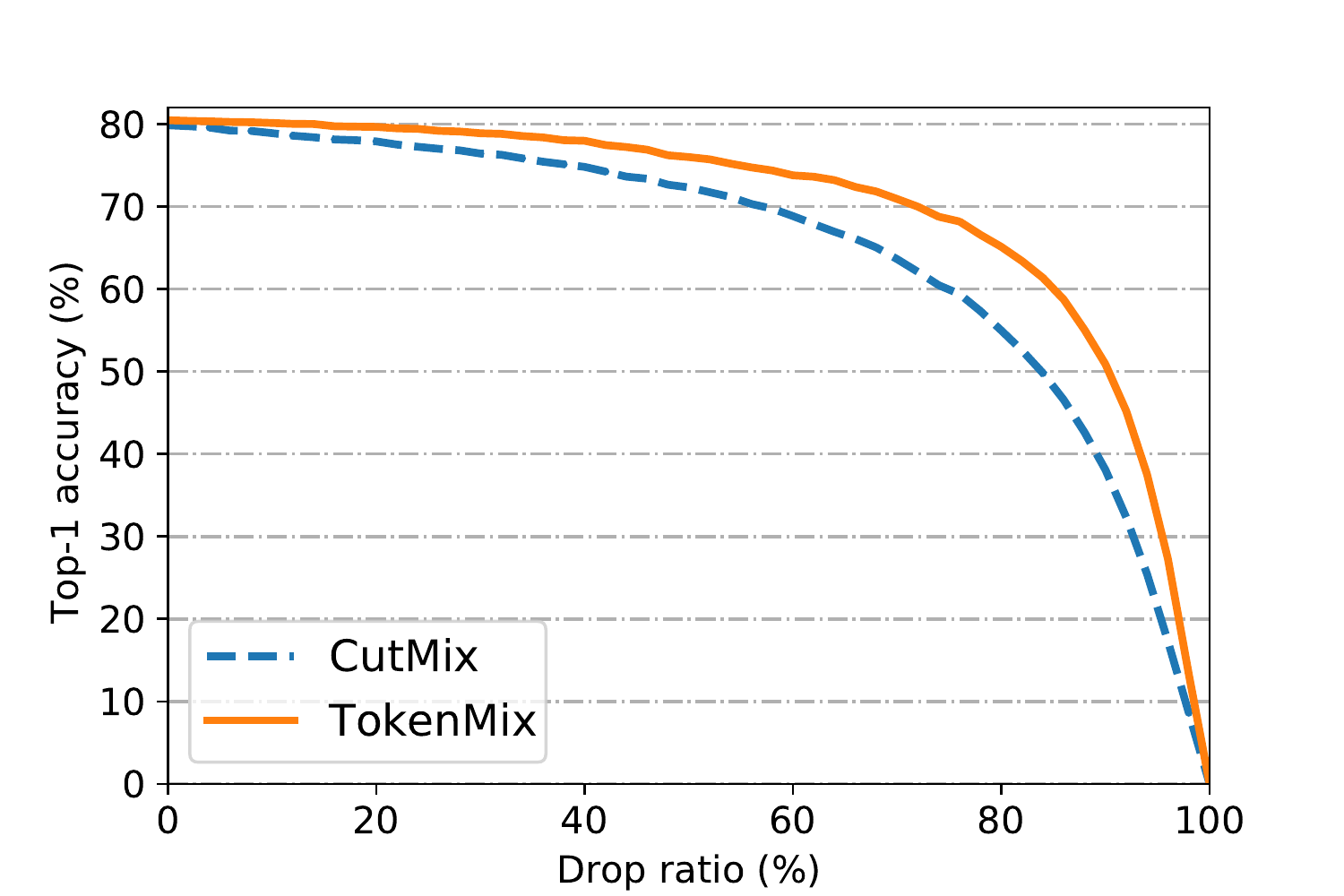}
    \caption{ImageNet top-1 accuracy of DeiT-S under different drop ratios. The gap between the model trained with CutMix and that with TokenMix increases as the drop ratio grows.}
    \label{fig:drop_score}
\end{wrapfigure}

\noindent\textbf{TokenMix enhances the occlusion robustness of vision transformers.}
After training converges, we construct a sequence of images with different occlusion ratios. Specifically, we gradually drop 10\% more patches and set the pixels inside to zero, and use the images for testing. We report the top-1 accuracy on ImageNet under different drop ratios. As shown in Figure \ref{fig:drop_score}, the model trained with TokenMix surpasses that with CutMix by increasingly larger margins as the drop ratio grows, demonstrating its better occlusion robustness. Specifically, we notice a $\sim$10\% performance gap at the drop ratio of 80\%.
We further visualize some examples in Figure  \ref{fig:occlusion_robustness}. It can be found that when about 40\% tokens are dropped, the predicted ground-truth class confidences of the baseline model decrease to very low values (0.25 in the first row) while the model trained with our TokenMix holds higher confidences (0.95 in the first row).

\begin{figure*}[t]
    \centering
    \includegraphics[width=0.75\linewidth]{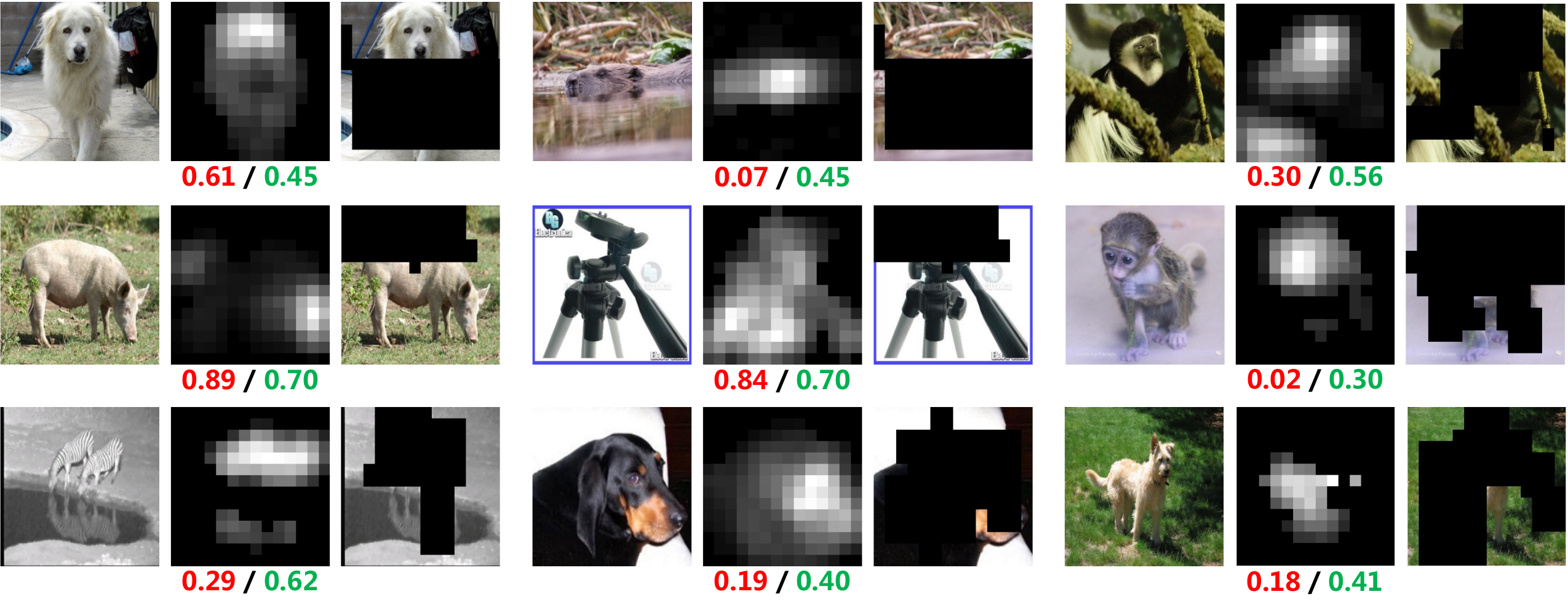}
    \caption{The target scores generated by \textcolor{red}{TokenMix} and \textcolor{blue}{CutMix}. For each tripled sub-figure, the left is the input image, the middle is the neural activation map, and the right is the masked image. Our approach generates more reasonable target scores, especially when the foreground region is cropped.}
    \label{fig:refine}
\end{figure*}

\section{Ablative studies}
\label{sec:ablation}

In this section, we conduct various ablation studies to analyze our proposed TokenMix. We use DeiT-S as the backbone and train it on ImageNet for 300 epochs unless otherwise specified. All other training settings are the same as described in Section \ref{sec:experiment}. We report the top-1 accuracy on ImageNet.


\begin{wraptable}{r}{0.45\textwidth}
  \centering
  \caption{Performances of using a single or randomly sampled one of the multiple mixing methods for training DeiT-B.}
  \resizebox{0.9\linewidth}{!}{
    \begin{tabular}{ccc|c}
    \toprule
    Mixup \cite{zhang2017mixup} & CutMix \cite{cutmix} & TokenMix & Top-1 Acc. \\
    \midrule
    \textcolor{mygray}{\ding{55}} & \textcolor{mygray}{\ding{55}} & \textcolor{mygray}{\ding{55}} & 75.8 \\
    \textcolor{mygray}{\ding{55}} & \ding{51} & \textcolor{mygray}{\ding{55}} & 78.7 \\
    \ding{51} & \textcolor{mygray}{\ding{55}} & \textcolor{mygray}{\ding{55}} & 80.0 \\
    \textcolor{mygray}{\ding{55}} & \textcolor{mygray}{\ding{55}} & \ding{51} & \textbf{81.5} \\
    \midrule
    \ding{51} & \ding{51} & \textcolor{mygray}{\ding{55}} & 81.8 \\
    \textcolor{mygray}{\ding{55}} & \ding{51} & \ding{51} & 82.0 \\
    \ding{51} & \textcolor{mygray}{\ding{55}} & \ding{51} & \textbf{82.9} \\
    \bottomrule
    \end{tabular}%
    }
  \label{tab:onlyonemix}%
\end{wraptable}

\noindent\textbf{Integrating TokenMix with previous mixing-based methods.}
Table \ref{tab:onlyonemix} presents the results of combining TokenMix with other mixing-based methods to train DeiT-B. When two mixing augmentations are utilized during training, one of them is randomly chosen to be used for data augmentation with a probability of $0.5$ at each iteration. The baseline (row 1 in Table \ref{tab:onlyonemix}) does not use any mixing-based augmentations. Using only MixToken improves the baseline and CutMix by large margins. Specifically, TokenMix improves the baseline by +5.7\% top-1 accuracy. Using both TokenMix and Mixup improves TokenMix-only by +1.4\% top-1 accuracy. In contrast, using both CutMix and Mixup also improves top-1 accuracy to 81.8\%, which, however, is still lower than TokenMix + Mixup.

\begin{table}[t]
    \centering
    \begin{minipage}[t]{0.4\linewidth}
    \captionof{table}{Comparison of different ways to generate neural activation maps. NFNet-F6 is used by default.}
    \centering
    \resizebox{0.9\linewidth}{!}{
        \begin{tabular}{lc|c}
        \toprule
        \makecell{Teacher} & \makecell{Teacher \\ Top-1 Acc.} & Top-1 Acc. \\
        \midrule
        NFNet-F6 \cite{nfnet} & 86.1 & 80.8 \\
        ResNet101 \cite{resnet}  & 82.3 & 80.7 \\
        ResNet26 \cite{resnet} & 79.8 & 80.5 \\
        Saliency \cite{saliency}  & N/A  & 80.1 \\
        \bottomrule
        \end{tabular}%
    }
    \label{tab:teacher}%
    \end{minipage}
    \centering
    \begin{minipage}[t]{0.53\linewidth}
        \centering
        \captionof{table}{Ablation of the way to generate target scores. Our approach generates target scores with content-based neural activation maps (denoted as \textit{refinement}).}
        \resizebox{0.6\linewidth}{!}{
        \begin{tabular}{ccc}
        \toprule
        Model & Refinement & Top-1 Acc. \\
        \midrule
        \multirow{2}{*}{DeiT-S \cite{deit}} & \textcolor{mygray}{\ding{55}} & 79.8 \\
         & \ding{51} & \textbf{80.5 \textcolor{blue}{(+0.7)}} \\
        \midrule
        \multirow{2}{*}{Swin-T \cite{swin}} & \textcolor{mygray}{\ding{55}} & 81.2 \\
         & \ding{51} & \textbf{81.5 \textcolor{blue}{(+0.3)}} \\
        \midrule
        \multirow{2}{*}{ResNet50 \cite{resnet}} & \textcolor{mygray}{\ding{55}} & 79.3 \\
         & \ding{51} & \textbf{79.8 \textcolor{blue}{(+0.5)}} \\
        \bottomrule
        \end{tabular}%
    }
    \label{tab:refine}%
    \end{minipage}
\end{table}

\begin{table}[t]
    
    \centering
    \begin{minipage}[t]{0.45\linewidth}
    \captionof{table}{Ablation of mask sampling strategy. The \textit{region-based} strategy works best on ResNet50, but degrades on DeiT-S.}
    \centering
    \resizebox{0.8\linewidth}{!}{
        \begin{tabular}{cccc}
        \toprule
        Model & region & random & block \\
        \midrule
        DeiT-T \cite{deit} & 72.2 & \textbf{72.7} & \textbf{72.7} \\
        DeiT-S \cite{deit} & 79.8 & \textbf{80.6} & \textbf{80.6} \\
        ResNet50 \cite{resnet} & 79.3 & 78.3 & \textbf{79.7} \\
        \bottomrule
        \end{tabular}%
        }
    \label{tab:masktype}%
    \end{minipage}
    \centering
    \begin{minipage}[t]{0.45\linewidth}
        \centering
        \captionof{figure}{Illustration of different mask sampling strategies.}
        \includegraphics[width=0.9\linewidth]{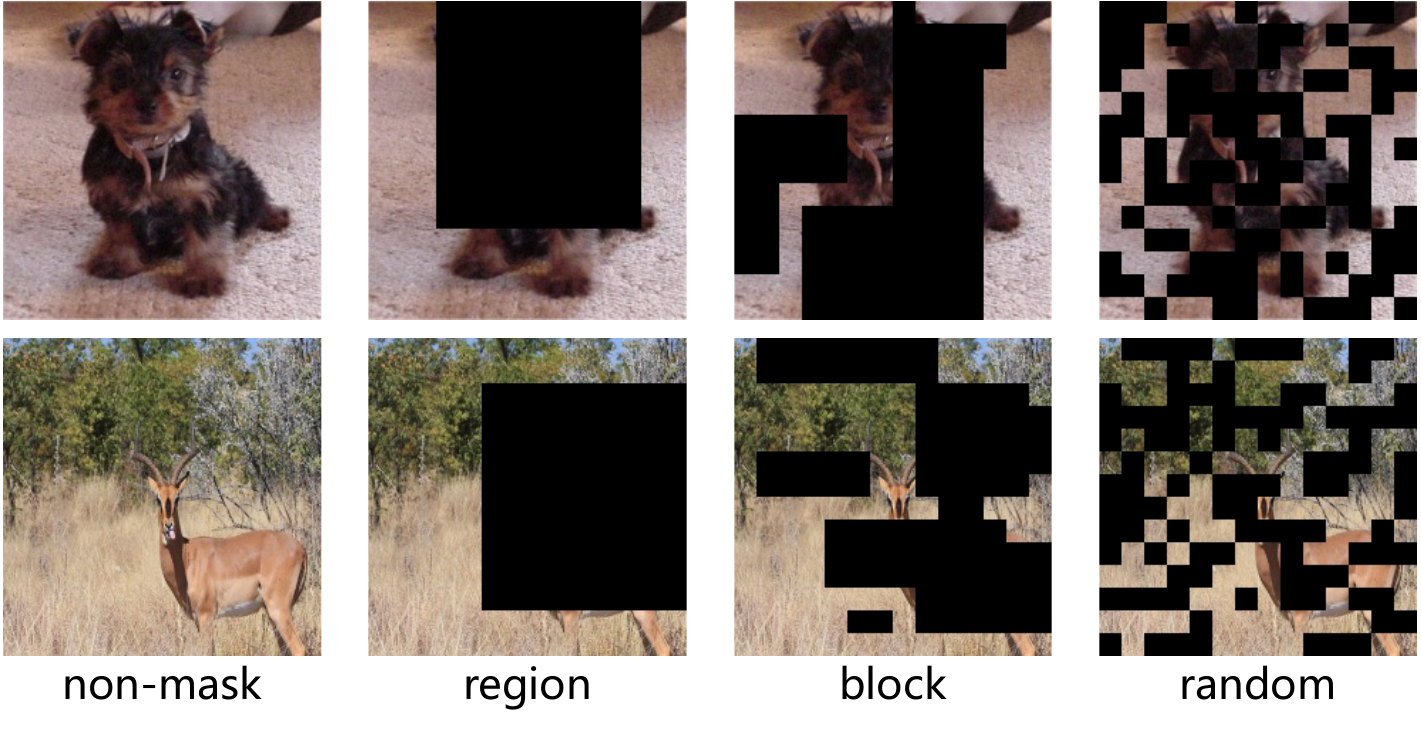}
        \label{fig:mask_type}
    \end{minipage}
\end{table}

\noindent\textbf{Different ways to generate neural activation maps.}
Table \ref{tab:teacher} presents the results of using different ways to generate neural activation maps. Besides our default choice of NFNet-F6, popular ResNet and hand-crafted saliency method are also compared. As shown in Table \ref{tab:teacher}, TokenMix follows the general behavior of distillation techniques, i.e., if the performance of the teacher model is high, so is the student model. 
In addition, TokenMix is robust to different choices of teacher networks for generating target scores. Even when the performance of the teacher drops by 6.3\%, only 0.3\% performance drop is observed on DeiT trained with our proposed TokenMix. However, using neural networks to generate neural activation maps is consistently better than using scores from hand-crafted approaches \cite{saliency} as the targets generated by a teacher model are generally better to learn than hand-crafted ones. We further visualized the target scores of the mixed images in Figure \ref{fig:refine}. For each tripled sub-figure, the left is the input images, the middle is the neural activation maps, and the right is the masked images. The scores generated from CutMix are shown in green, while the red scores are generated by our approach. As shown in  Figure \ref{fig:refine}, the target scores generated by our approach are more reasonable, especially when the foreground is cut.

\noindent\textbf{Our neural-activation target scores are comptible with CutMix.}
To test whether our proposed target scores are compatible with CutMix, we conduct experiments of using CutMix to mix pairs of images but generating targets with our approach (denoted as \textit{refinement} in Table~\ref{tab:refine}), and train with various backbones, e.g., DeiT-S, Swin-T, and ResNet50. As shown in Table~\ref{tab:refine}, we achieve consistent performance gains on those backbones. Specifically, we improve DeiT-S for +0.7\%, Swin-T for +0.3\%, and ResNet50 for +0.5\% with nearly no extra computation cost during the training process.
All the results verify the compatibility of our proposed target score assignment with CutMix.
\begin{wraptable}{r}{0.4\textwidth}
  \centering
  \caption{Ablation of mask sampling strategy. The \textit{block-based} strategy obtains higher accuracy with label refinement.}
  \resizebox{0.8\linewidth}{!}{
    \begin{tabular}{c|ccc}
    \toprule
    Model & Mask & Refinement & Top-1 Acc. \\
    \midrule
    
    \multirow{4}{*}{DeiT-T} & \multirow{2}{*}{random} & \textcolor{mygray}{\ding{55}} & 72.7 \\
     & & \ding{51} & \textbf{72.9 \textcolor{blue}{(+0.2)}} \\
    \cmidrule{2-4}
     & \multirow{2}{*}{block} & \textcolor{mygray}{\ding{55}} & 72.7 \\
     & & \ding{51} & \textbf{73.2 \textcolor{blue}{(+0.5)}} \\
    \midrule
    
    \multirow{4}{*}{DeiT-S} & \multirow{2}{*}{random} & \textcolor{mygray}{\ding{55}} & 80.6 \\
     & & \ding{51} & 80.6 \\
    \cmidrule{2-4}
     & \multirow{2}{*}{block} & \textcolor{mygray}{\ding{55}} & 80.6 \\
     & & \ding{51} & \textbf{80.8 \textcolor{blue}{(+0.2)}} \\
    \bottomrule
    \end{tabular}%
    }
  \label{tab:masktype_refine}%
\end{wraptable}

\noindent\textbf{Mask sampling strategy.} Table \ref{tab:masktype} presents the impact of different sampling strategies on transformers and Convolution Neural Networks, as illustrated in Figure \ref{fig:mask_type}. The \textit{region-based} sampling, widely used in \cite{cutmix,saliency,attentivemix}, cuts a single large rectangular area from the mixing images. Our proposed TokenMix directly cuts at token-level. We compare two settings of our approach, masking multiple blocks (\textit{block-based}) following our description in Section \ref{sec:tokenmix} or masking individual tokens separately (\textit{random}). To better inspect the impact of sampling strategies alone, in all the experiments in Table \ref{tab:masktype}, we directly use target scores generated by CutMix, instead of ours.
As shown in Table \ref{tab:masktype}, the \textit{region-based} strategy achieves decent performance on ResNet50, but degrades on transformers, which validates our argument on the sub-optimality of using region-based cut for training transformers.
Compared to the \textit{random} strategy, the \textit{block-based} strategy achieves similar performances on transformers, but performs much better on ResNet50.
When using our proposed target score assignment approach, we further compare the \textit{random} and the \textit{block-based} sampling strategies. As shown in Table \ref{tab:masktype_refine}, the \textit{block-based} strategy used in our final solution consistently has higher accuracy. The results further verify the effectiveness of our proposed TokenMix.

\begin{table}[t]
    \centering
    \begin{minipage}[t]{0.45\linewidth}
    \captionof{table}{Ablation of training epochs. TokenMix enjoys longer training. The extra 100 epochs of training improve +0.4\% accuracy.}
    \centering
    \resizebox{0.75\linewidth}{!}{
        \begin{tabular}{ccc}
        \toprule
        Mixing Method & Epoch & Top-1 Acc. \\
        \midrule
        \multirow{2}{*}{CutMix \cite{cutmix}} & 300 & 79.8 \\
        & 400 & 79.9 \textcolor{blue}{(+0.1)} \\
        \midrule
        \multirow{2}{*}{TokenMix} & 300 & 80.8 \\
        & 400 & \textbf{81.2 \textcolor{blue}{(+0.4)}} \\
        \bottomrule
        \end{tabular}%
    }
    \label{tab:epoch}%
    \end{minipage}
    \centering
    \begin{minipage}[t]{0.45\linewidth}
        \centering
        \captionof{table}{Ablation of the loss function. Binary cross-entropy (BCE) improves TokenMix, compared with multi-class cross-entropy (CE).}
        \resizebox{0.85\linewidth}{!}{
            \begin{tabular}{ccc}
            \toprule
            Mixing Method & Loss Type & Top-1 Acc. \\
            \midrule
            \multirow{2}{*}{CutMix \cite{cutmix}} & CE & 79.8 \\
             & BCE & 79.8 \\
            \midrule
            \multirow{2}{*}{TokenMix} & CE & 80.3 \\
             & BCE & \textbf{80.8 \textcolor{blue}{(+0.5)}} \\
            \bottomrule
            \end{tabular}
            }
        \label{tab:loss}
    \end{minipage}
\end{table}

\noindent\textbf{Training epochs.} Table \ref{tab:epoch} presents the results of longer training. As targets generated by a teacher network's neural activation maps can provide more appropriate scores and more challenging samples for training the transformers, which mitigates the risk of over-fitting scheme, our proposed TokenMix can enjoy longer training. As shown in Table \ref{tab:epoch}, our TokenMix improves DeiT-S for +0.4\% with additional 100 training epochs, while using CutMix for long training is less beneficial.

\noindent\textbf{Loss function.}
As the mixed images may contain multiple objects of different classes, we adopt the binary cross-entropy (BCE) loss instead of the typical cross-entropy (CE) loss \cite{timmresnet,imagenetdone}. Using BCE loss improves DeiT-S for +0.5\% accuracy (Table \ref{tab:loss}) when training with our proposed TokenMix, as the cut-and-paste operation might generate a mixed image with multiple objects of different classes.
It might be because the generated targets by CutMix are sub-optimal, we do not notice performance improvement of replacing CE with BCE when training DeiT-S with CutMix augmentation.

\section{Conclusions}
\label{sec:conclusion}
In this paper, we propose TokenMix, a token-level augmentation strategy that generalizes well across various transformer-based architectures. TokenMix is motivated by two key observations: 1) region-level mixing is less beneficial for transformer-based architectures, and 2) assigning target of mixed images with linear combination might be inaccurate and even counter-intuitive. Our proposed TokenMix directly cuts at token level and obtains the target of the mixed images with content-based neural activation maps. Empirical results show that TokenMix has the properties of enhancing occlusion robustness and helping vision transformers focus on the foreground area of input images. Besides, TokenMix consistently improves various transformer-based architectures, including DeiT, PVT, and Swin Transformer.

\textbf{Acknowledgement}
Hongsheng Li is also a Principal Investigator of Centre for Perceptual and Interactive Intelligence Limited (CPII). This work is supported in part by CPII, in part by the General Research Fund through the Research Grants Council of Hong Kong under Grants (Nos. 14204021, 14207319), in part by CUHK Strategic Fund.

\section{Appendix}

\subsection{Training Details}
\subsubsection{ImageNet-1K}

Our training receipt follows common practice of supervised ViT training. The default setting is in Table \ref{tab:traing_setting}. 

\begin{table}[h]
    \centering
    \caption{Training settings on ImageNet-1K. $\ast$ optional config.}
    \begin{tabular}{l|c}
        \toprule
        config & value \\
        \midrule
        optimizer & AdamW \cite{adamw} \\
        learning rate & 0.001 \\
        weight decay & 0.05 \\
        batch size & 1024 \\
        learning rate schedule & cosine decay \\
        warmup epochs & 5 \\
        training epochs & 300 \\
        augmentation & RandAug(9, 0.5) \cite{randaug} \\
        LabelSmooth \cite{inception} & 0.1 \\
        DropPath \cite{droppath} & 0.1 \\
        Mixup \textsuperscript{$\ast$} \cite{zhang2017mixup} & 0.8 \\
        CutMix \textsuperscript{$\ast$} \cite{cutmix} & 1.0 \\
        TokenMix \textsuperscript{$\ast$} & 0.5 \\
        \bottomrule
    \end{tabular}
    \label{tab:traing_setting}
\end{table}

\subsubsection{ADE20K}
For transferring to ADE20K, we directly follow the task layer and most of the hyperparameters described in SETR-PUP \cite{setr}. The detailed hyperparameters are described in Table \ref{tab:ade20k}.

\begin{table}[h]
    \centering
    \caption{Training settings on ADE20K.}
    \begin{tabular}{l|c}
        \toprule
        config & value \\
        \midrule
        optimizer & Adam \cite{adam} \\
        learning rate & 0.001 \\
        weight decay & 0.05 \\
        batch size & 16 \\
        learning rate schedule & linear \\
        warmup steps & 1500 \\
        training steps & 160K \\
        input resolution & 512 $\times$ 512 \\
        \bottomrule
    \end{tabular}
    \label{tab:ade20k}
\end{table}

\subsection{ImageNet Examples}
We visualize some examples in ImageNet-1K (Figure \ref{fig:example}) to show the disagreement between the salient areas and the foreground regions of the target class.
\begin{figure}[h]
    \centering
    \includegraphics[width=0.9\linewidth]{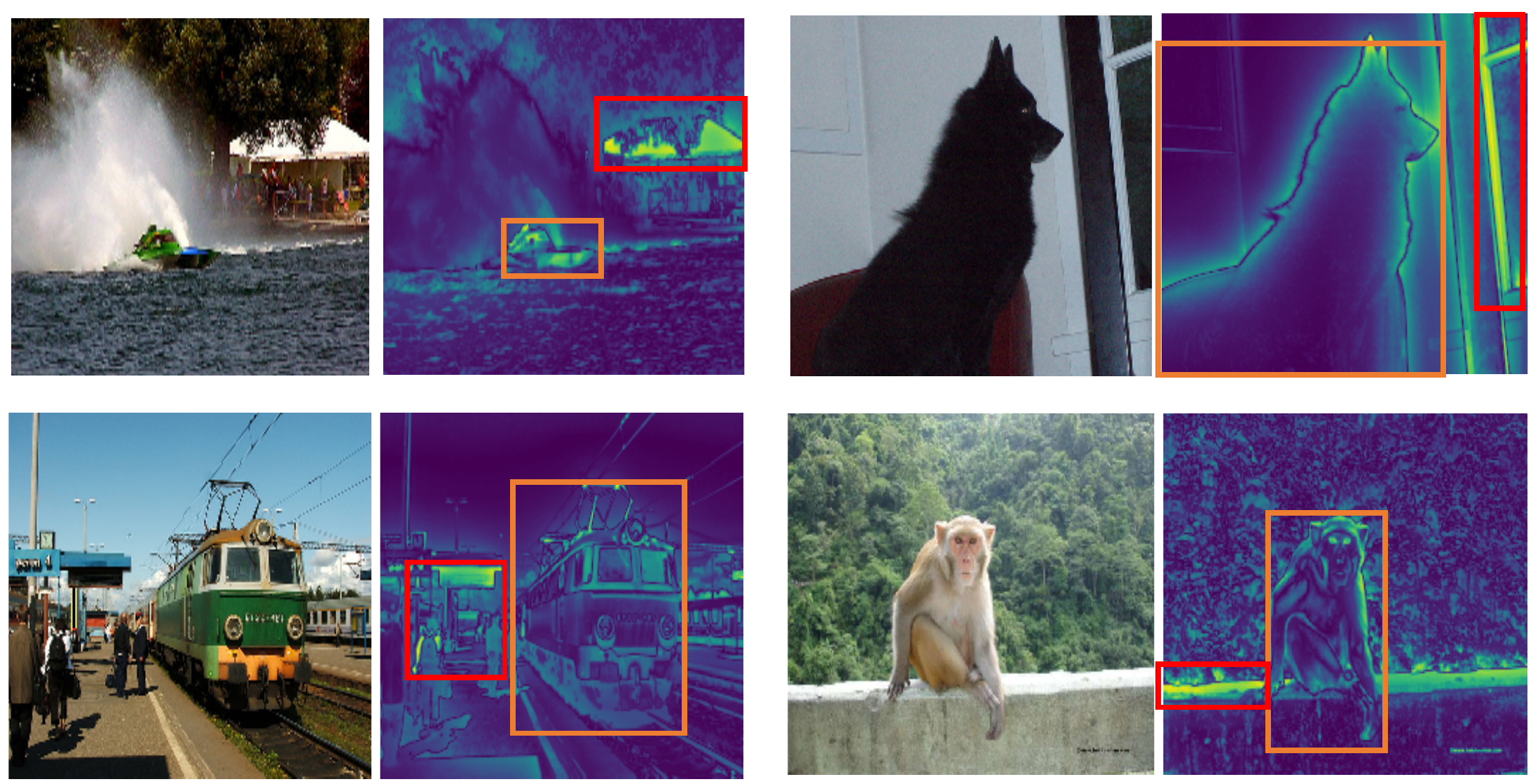}
    \caption{Examples from ImageNet-1K. \textcolor{orange}{Orange boxes} indicate foreground regions of the target classes. \textcolor{red}{Red boxes} indicate the most salient areas.}
    \label{fig:example}
\end{figure}

\clearpage
%
%
\bibliographystyle{splncs04}
\bibliography{egbib}

\begin{thebibliography}{10}
\providecommand{\url}[1]{\texttt{#1}}
\providecommand{\urlprefix}{URL }
\providecommand{\doi}[1]{https://doi.org/#1}

\bibitem{beit}
Bao, H., Dong, L., Wei, F.: Beit: Bert pre-training of image transformers.
  arXiv preprint arXiv:2106.08254  (2021)

\bibitem{imagenetdone}
Beyer, L., H{\'e}naff, O.J., Kolesnikov, A., Zhai, X., Oord, A.v.d.: Are we
  done with imagenet? arXiv preprint arXiv:2006.07159  (2020)

\bibitem{nfnet}
Brock, A., De, S., Smith, S.L., Simonyan, K.: High-performance large-scale
  image recognition without normalization. arXiv preprint arXiv:2102.06171
  (2021)

\bibitem{vistransformer}
Chefer, H., Gur, S., Wolf, L.: Transformer interpretability beyond attention
  visualization. In: Proceedings of the IEEE/CVF Conference on Computer Vision
  and Pattern Recognition. pp. 782--791 (2021)

\bibitem{Chen2021TransMixAT}
Chen, J., Sun, S., He, J., Torr, P.H.S., Yuille, A.L., Bai, S.: Transmix:
  Attend to mix for vision transformers. ArXiv  \textbf{abs/2111.09833} (2021)

\bibitem{Chen2020GridMaskDA}
Chen, P., Liu, S., Zhao, H., Jia, J.: Gridmask data augmentation. ArXiv
  \textbf{abs/2001.04086} (2020)

\bibitem{autoaug}
Cubuk, E.D., Zoph, B., Mane, D., Vasudevan, V., Le, Q.V.: Autoaugment: Learning
  augmentation policies from data. arXiv preprint arXiv:1805.09501  (2018)

\bibitem{randaugment}
Cubuk, E.D., Zoph, B., Shlens, J., Le, Q.V.: Randaugment: Practical automated
  data augmentation with a reduced search space. In: Proceedings of the
  IEEE/CVF Conference on Computer Vision and Pattern Recognition Workshops. pp.
  702--703 (2020)

\bibitem{randaug}
Cubuk, E.D., Zoph, B., Shlens, J., Le, Q.V.: Randaugment: Practical automated
  data augmentation with a reduced search space. In: Proceedings of the
  IEEE/CVF Conference on Computer Vision and Pattern Recognition Workshops. pp.
  702--703 (2020)

\bibitem{imagenet}
Deng, J., Dong, W., Socher, R., Li, L.J., Li, K., Fei-Fei, L.: Imagenet: A
  large-scale hierarchical image database. In: 2009 IEEE conference on computer
  vision and pattern recognition. pp. 248--255. Ieee (2009)

\bibitem{Devries2017ImprovedRO}
Devries, T., Taylor, G.W.: Improved regularization of convolutional neural
  networks with cutout. ArXiv  \textbf{abs/1708.04552} (2017)

\bibitem{vit}
Dosovitskiy, A., Beyer, L., Kolesnikov, A., Weissenborn, D., Zhai, X.,
  Unterthiner, T., Dehghani, M., Minderer, M., Heigold, G., Gelly, S., et~al.:
  An image is worth 16x16 words: Transformers for image recognition at scale.
  arXiv preprint arXiv:2010.11929  (2020)

\bibitem{resnet}
He, K., Zhang, X., Ren, S., Sun, J.: Deep residual learning for image
  recognition. In: Proceedings of the IEEE conference on computer vision and
  pattern recognition. pp. 770--778 (2016)

\bibitem{Hendrycks2020AugMixAS}
Hendrycks, D., Mu, N., Cubuk, E.D., Zoph, B., Gilmer, J., Lakshminarayanan, B.:
  Augmix: A simple data processing method to improve robustness and
  uncertainty. ArXiv  \textbf{abs/1912.02781} (2020)

\bibitem{droppath}
Huang, G., Sun, Y., Liu, Z., Sedra, D., Weinberger, K.Q.: Deep networks with
  stochastic depth. In: European conference on computer vision. pp. 646--661.
  Springer (2016)

\bibitem{tokenlabeling}
Jiang, Z., Hou, Q., Yuan, L., Zhou, D., Jin, X., Wang, A., Feng, J.: Token
  labeling: Training a 85.5\% top-1 accuracy vision transformer with 56m
  parameters on imagenet. arXiv preprint arXiv:2104.10858  (2021)

\bibitem{comix}
Kim, J.H., Choo, W., Jeong, H., Song, H.O.: Co-mixup: Saliency guided joint
  mixup with supermodular diversity. arXiv preprint arXiv:2102.03065  (2021)

\bibitem{puzzlemix}
Kim, J.H., Choo, W., Song, H.O.: Puzzle mix: Exploiting saliency and local
  statistics for optimal mixup. In: International Conference on Machine
  Learning. pp. 5275--5285. PMLR (2020)

\bibitem{adam}
Kingma, D.P., Ba, J.: Adam: A method for stochastic optimization. arXiv
  preprint arXiv:1412.6980  (2014)

\bibitem{coco}
Lin, T.Y., Maire, M., Belongie, S., Hays, J., Perona, P., Ramanan, D.,
  Doll{\'a}r, P., Zitnick, C.L.: Microsoft coco: Common objects in context. In:
  European conference on computer vision. pp. 740--755. Springer (2014)

\bibitem{swin}
Liu, Z., Lin, Y., Cao, Y., Hu, H., Wei, Y., Zhang, Z., Lin, S., Guo, B.: Swin
  transformer: Hierarchical vision transformer using shifted windows. arXiv
  preprint arXiv:2103.14030  (2021)

\bibitem{adamw}
Loshchilov, I., Hutter, F.: Decoupled weight decay regularization. arXiv
  preprint arXiv:1711.05101  (2017)

\bibitem{saliency}
Montabone, S., Soto, A.: Human detection using a mobile platform and novel
  features derived from a visual saliency mechanism. Image and Vision Computing
   \textbf{28}(3),  391--402 (2010)

\bibitem{Qin2020ResizeMixMD}
Qin, J., Fang, J., Zhang, Q., Liu, W., gang Wang, X., Wang, X.: Resizemix:
  Mixing data with preserved object information and true labels. ArXiv
  \textbf{abs/2012.11101} (2020)

\bibitem{setr}
S, Z., J, L., H, Z., X, Z., Z, L., Y, W., Y, F., J, F., T, X., PH, T., L., Z.:
  Rethinking semantic segmentation from a sequence-to-sequence perspective with
  transformers. In: Proceedings of the IEEE/CVF conference on computer vision
  and pattern recognition 2021. pp. 6881--6890 (2020)

\bibitem{Singh2018HideandSeekAD}
Singh, K.K., Yu, H., Sarmasi, A., Pradeep, G., Lee, Y.J.: Hide-and-seek: A data
  augmentation technique for weakly-supervised localization and beyond. ArXiv
  \textbf{abs/1811.02545} (2018)

\bibitem{inception}
Szegedy, C., Vanhoucke, V., Ioffe, S., Shlens, J., Wojna, Z.: Rethinking the
  inception architecture for computer vision. In: Proceedings of the IEEE
  conference on computer vision and pattern recognition. pp. 2818--2826 (2016)

\bibitem{Takahashi2020DataAU}
Takahashi, R., Matsubara, T., Uehara, K.: Data augmentation using random image
  cropping and patching for deep cnns. IEEE Transactions on Circuits and
  Systems for Video Technology  \textbf{30},  2917--2931 (2020)

\bibitem{deit}
Touvron, H., Cord, M., Douze, M., Massa, F., Sablayrolles, A., J{\'e}gou, H.:
  Training data-efficient image transformers \& distillation through attention.
  In: International Conference on Machine Learning. pp. 10347--10357. PMLR
  (2021)

\bibitem{cait}
Touvron, H., Cord, M., Sablayrolles, A., Synnaeve, G., J{\'e}gou, H.: Going
  deeper with image transformers. arXiv preprint arXiv:2103.17239  (2021)

\bibitem{saliencymix}
Uddin, A., Monira, M., Shin, W., Chung, T., Bae, S.H., et~al.: Saliencymix: A
  saliency guided data augmentation strategy for better regularization. arXiv
  preprint arXiv:2006.01791  (2020)

\bibitem{alignmixup}
Venkataramanan, S., Kijak, E., Amsaleg, L., Avrithis, Y.: Alignmixup: Improving
  representations by interpolating aligned features. In: Proceedings of the
  IEEE/CVF Conference on Computer Vision and Pattern Recognition. pp.
  19174--19183 (2022)

\bibitem{Verma2019ManifoldMB}
Verma, V., Lamb, A., Beckham, C., Najafi, A., Mitliagkas, I., Lopez-Paz, D.,
  Bengio, Y.: Manifold mixup: Better representations by interpolating hidden
  states. In: ICML (2019)

\bibitem{attentivemix}
Walawalkar, D., Shen, Z., Liu, Z., Savvides, M.: Attentive cutmix: An enhanced
  data augmentation approach for deep learning based image classification.
  arXiv preprint arXiv:2003.13048  (2020)

\bibitem{pvt}
Wang, W., Xie, E., Li, X., Fan, D.P., Song, K., Liang, D., Lu, T., Luo, P.,
  Shao, L.: Pyramid vision transformer: A versatile backbone for dense
  prediction without convolutions. arXiv preprint arXiv:2102.12122  (2021)

\bibitem{timmresnet}
Wightman, R., Touvron, H., J{\'e}gou, H.: Resnet strikes back: An improved
  training procedure in timm. arXiv preprint arXiv:2110.00476  (2021)

\bibitem{cutmix}
Yun, S., Han, D., Oh, S.J., Chun, S., Choe, J., Yoo, Y.: Cutmix: Regularization
  strategy to train strong classifiers with localizable features. In:
  Proceedings of the IEEE/CVF International Conference on Computer Vision. pp.
  6023--6032 (2019)

\bibitem{relabel}
Yun, S., Oh, S.J., Heo, B., Han, D., Choe, J., Chun, S.: Re-labeling imagenet:
  from single to multi-labels, from global to localized labels. In: Proceedings
  of the IEEE/CVF Conference on Computer Vision and Pattern Recognition. pp.
  2340--2350 (2021)

\bibitem{zhang2017mixup}
Zhang, H., Cisse, M., Dauphin, Y.N., Lopez-Paz, D.: mixup: Beyond empirical
  risk minimization. arXiv preprint arXiv:1710.09412  (2017)

\bibitem{Zhong2020RandomED}
Zhong, Z., Zheng, L., Kang, G., Li, S., Yang, Y.: Random erasing data
  augmentation. ArXiv  \textbf{abs/1708.04896} (2020)

\bibitem{cam}
Zhou, B., Khosla, A., Lapedriza, A., Oliva, A., Torralba, A.: Learning deep
  features for discriminative localization. In: Proceedings of the IEEE
  conference on computer vision and pattern recognition. pp. 2921--2929 (2016)

\bibitem{ade20k}
Zhou, B., Zhao, H., Puig, X., Fidler, S., Barriuso, A., Torralba, A.: Scene
  parsing through ade20k dataset. In: Proceedings of the IEEE conference on
  computer vision and pattern recognition. pp. 633--641 (2017)

\end{thebibliography}
\end{document}